\documentclass[10pt,twocolumn,letterpaper]{article}

\usepackage[pagenumbers]{wacv} % To force page numbers, e.g. for an arXiv version

\usepackage{siunitx}
\usepackage{tabularx}
\usepackage{multirow}
\usepackage{pifont}
\usepackage{tikz}

\newcommand{\cmark}{\ding{51}}%
\newcommand{\xmark}{\ding{55}}%

\newcommand\blfootnote[1]{%
\begingroup
\renewcommand\thefootnote{}{}\footnote{#1}%
\addtocounter{footnote}{-1}%
\endgroup
}

\definecolor{wacvblue}{rgb}{0.21,0.49,0.74}
\usepackage[pagebackref,breaklinks,colorlinks,allcolors=wacvblue]{hyperref}

\title{NavMapFusion: Diffusion-based Fusion of Navigation Maps\\
for Online Vectorized HD Map Construction}

\author{Thomas Monninger\textsuperscript{1,2} \quad Zihan Zhang\textsuperscript{*,3} \quad Steffen Staab\textsuperscript{2,4} \quad Sihao Ding\textsuperscript{1}
\vspace{3mm}
\\
\small{\textsuperscript{1}Mercedes-Benz Research \& Development North America, USA} \\%~~
\small{\textsuperscript{2}University of Stuttgart, Germany} \\%~~
\small{\textsuperscript{3}University of California, San Diego, USA} \\%~~
\small{\textsuperscript{4}University of Southampton, United Kingdom
}
}

\newcommand\copyrighttext{\footnotesize \textcopyright~2025 IEEE. Personal use of this material is permitted.  Permission from IEEE must be obtained for all other uses, in any current or future media, including reprinting/republishing this material for advertising or promotional purposes, creating new collective works, for resale or redistribution to servers or lists, or reuse of any copyrighted component of this work in other works.
}%

\newcommand\copyrightnotice{%
    \begin{tikzpicture}[remember picture,overlay]%
     \node[anchor=south, xshift=0pt, yshift=12pt] at (current page.south)%
     {\fbox{\parbox{\dimexpr\textwidth-\fboxsep-\fboxrule\relax}{\copyrighttext}}};%
     \end{tikzpicture}%
}

\begin{document}
\maketitle

\blfootnote{$(*)$ Work was done during an internship at Mercedes-Benz Research \&
Development North America.}

\vspace{-0.3cm}
\begin{abstract}
Accurate environmental representations are essential for autonomous driving, providing the foundation for safe and efficient navigation.
Traditionally, high‑definition (HD) maps are providing this representation of the static road infrastructure to the autonomous system a priori.
However, because the real world is constantly changing, such maps must be constructed online from on‑board sensor data.  
Navigation‑grade standard-definition (SD) maps are widely available, but their resolution is insufficient for direct deployment. Instead, they can be used as coarse prior to guide the online map construction process.  
We propose \textbf{NavMapFusion}, a diffusion‑based framework that performs iterative denoising conditioned on high-fidelity sensor data and on low-fidelity navigation maps.  
This paper strives to answer:  
(1) How can coarse, potentially outdated navigation maps guide online map construction?  
(2) What advantages do diffusion models offer for map fusion?  
We demonstrate that diffusion-based map construction provides a robust framework for map fusion. 
Our key insight is that discrepancies between the prior map and online perception naturally correspond to noise within the diffusion process; consistent regions reinforce the map construction, whereas outdated segments are suppressed.
On the nuScenes benchmark, NavMapFusion conditioned on coarse road lines from OpenStreetMap data reaches a 21.4~\% relative improvement on 100~m, and even stronger improvements on larger perception ranges, while maintaining real‑time capabilities.  
By fusing low‑fidelity priors with high‑fidelity sensor data, the proposed method generates accurate and up-to-date environment representations, guiding towards safer and more reliable autonomous driving.
The code is available at \href{https://github.com/tmonnin/navmapfusion}{https://github.com/tmonnin/navmapfusion}.
\end{abstract}

\vspace{-0.2cm}
\begin{figure}[t!]
    \centering
    \includegraphics[width=1.0\columnwidth]{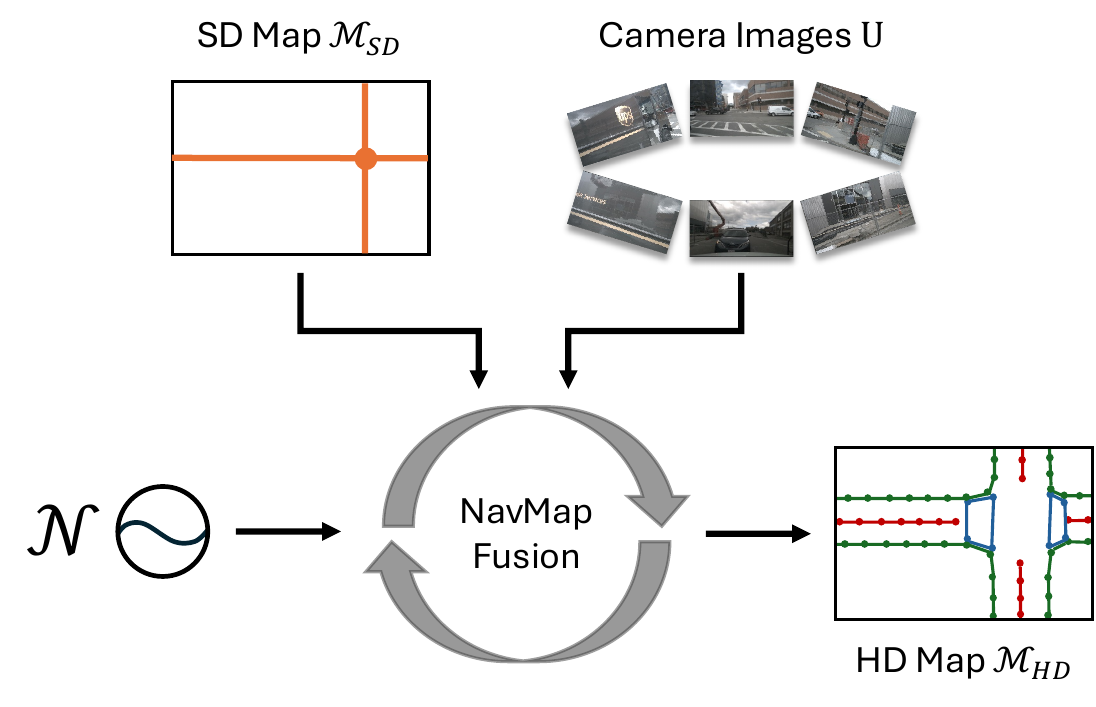}
    \caption{Overview of our NavMapFusion approach. Diffusion-based map construction starts from random noise and is conditioned on camera images and SD map to generate an HD map.}
    \label{fig:navmapfusion_overview}
\end{figure}

\section{Introduction}
Accurate knowledge of static road infrastructure, such as lanes, dividers, and crosswalks, is essential for decision making in autonomous vehicles.  
This knowledge must be extracted from sensor data to react to the actual environment around the vehicle in real time.
However, limited range and occlusions impose limits to pure perception-based online mapping.  
Navigation maps offer complementary global context but lack the resolution and might be outdated; consequently, they can be used as \emph{guidance} \cite{luo2024smerf, schmidt2023exploring}.
Leveraging coarse priors during online HD‑map construction can close perception gaps in occluded or far‑distance regions, improving safety margins and planning performance.
{\copyrightnotice}

Conflicts between the navigation prior and online sensor observations may stem from true environment changes (\eg, construction) or from limited sensor view (\eg, occlusion).
A fusion algorithm must therefore perform context‑aware reasoning: retaining correct but currently invisible structures while discarding obsolete ones.  
This is particularly challenging since prior maps are mostly fully correct, but sometimes locally wrong due to roadwork. 
Another source of error is inaccurate localization, causing systematic errors, drifts, or sudden jumps.
The non‑uniform spatial error profile of real‑world maps renders heuristics-based map fusion unreliable.

Classical late‑stage fusion pipelines treat perception output and prior maps as separate layers, deferring a hard decision until the end; this struggles when the inputs disagree.  
Recent learning‑based approaches use neural network architectures to condition the online map construction with prior map information. 
Their deterministic fusion process makes it harder to discard stale information.  
In contrast, we embed the conditioning inside a diffusion framework, allowing the model to attenuate or amplify individual elements in a probabilistic manner.
Experiments on nuScenes confirm that integrating prior maps through a diffusion process is effective for map fusion and outperforms state‑of‑the‑art baselines.

In summary, our contributions are: (1) we propose NavMapFusion, a novel framework that leverages a diffusion process to fuse navigation map priors with sensor data for online HD map construction; (2) we demonstrate that diffusion-based map fusion is more effective than deterministic fusion through experiments on the nuScenes dataset; (3) we provide an extensive study on the robustness of NavMapFusion towards errors in the SD map input.

\section{Related Work}\label{sec:related_work}

\subsection{Online Map Construction}
Philion and Fidler \cite{liftsplatshoot_2020} propose the first learning-based architecture for raster map construction in an online setup.
BEVFormer \cite{monninger2024tempbev} and TempBEV \cite{bevformer_2022} improve accuracy by aggregating temporal information across multiple time steps.
BEVerse \cite{beverse_2022} and BEVSegformer \cite{bevsegformer_2023} achieve further improvements on constructing a raster map.
Li \etal \cite{hdmapnet_2022} perform map segmentation first and add a post-processing step that outputs vectorized map geometries.

Liu \etal present VectorMapNet \cite{vectormapnet_2023}, the first end-to-end model for vectorized map learning.
Further, MapTR \cite{maptr_2023} addresses the ambiguity in selecting a discrete set of points to model geometries in vectorized representations by employing permutation-equivalent modeling, which stabilizes the learning process.
Zhang \etal \cite{zhang2024online} propose a geometric loss function that is robust to rigid transformations.
StreamMapNet \cite{streammapnet_2024} and MapUnveiler \cite{kim2024unveiling} are more recent approaches that address temporal stability in constructed online maps.
AugMapNet \cite{monninger2025augmapnet} improves the spatial structure of the latent space with dense spatial supervision.
SuperFusion \cite{dong2024superfusion} and ScalableMap \cite{yu2023scalablemap} address long-range online map construction without the use of prior maps.

\begin{figure}[tbp]
    \centering
    \includegraphics[width=\columnwidth]{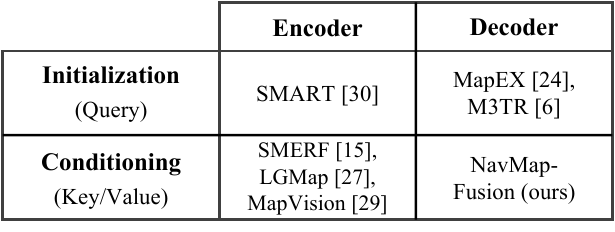}
    \vspace{-6mm}
    \caption{Categorization of related work on prior map fusion for transformer-based vectorized online map construction.}
    \label{fig:navmap_diffusion_related_work}
    \vspace{-1mm}
\end{figure}

\subsection{Diffusion-based Map Construction}
Recent work explored the use of diffusion models for online raster map generation from on-road camera inputs. DiffMap \cite{jia2024diffmap} introduces a latent diffusion model that improves raster map quality by incorporating structured priors from segmentation masks. DifFUSER \cite{le2024diffusion} extends diffusion models to handle both 3D object detection and rasterized map prediction.
In contrast to raster map approaches, our work focuses on generating vector-based map elements using diffusion. 
PolyDiffuse \cite{Chen2023PolyDiffuse} utilizes diffusion for online vectorized map generation. Its Guided Set Diffusion Model refines coarse map predictions from existing models. 
In contrast, MapDiffusion \cite{monninger2025mapdiffusion} fully formulates vectorized online map construction as a generative diffusion process, starting from random noise without relying on coarse initializations. 
Hence, MapDiffusion is the foundation for our approach.

\subsection{Map Construction with Navigation Map Priors}

Navigation maps provide strong priors for online map construction. Raster-based approaches encode prior maps and integrate them via attention or convolution. P-MapNet~\cite{jiang2024pmapnet} and BLOS-BEV~\cite{wu2024blos} use cross-attention to fuse raster priors with sensor data. RoadPainter~\cite{ma2024roadpainter} renders priors into BEV features and applies self-/cross-attention in the decoder. EORN~\cite{zhang2024enhancing} updates BEV features with raster SD maps through convolution and concatenation. NeuralMapPrior~\cite{xiong2023neuralmapprior} extends BEV latent space using priors with attention and GRUs. CoGMP~\cite{fu2025generative} employs diffusion-based generation conditioned on structured vector elements.

Few works exist on guiding vectorized online map construction with information from prior maps, \cref{fig:navmap_diffusion_related_work} provides a schematic overview.
SMART \cite{ye2025smart} encodes navigation maps and satellite images into a BEV grid that substitutes the learnable BEV queries in the BEV encoder.
SMERF \cite{luo2024smerf}, LGMap \cite{wu2024lgmap}, and MapVision \cite{yang2024mapvision} encode the map prior into latent map features, which are integrated into the latent BEV grid by extending the encoder process with an additional map cross-attention step.
MapEx \cite{sun2023mind} and M3TR \cite{immel2024m3tr} encode the navigation map elements into queries that are used as a starting point for decoding the online map.
NavMapFusion is filling the white spot by performing conditioning on the decoder side.

\section{Approach}

\begin{figure}[tbp]
    \centering
    \includegraphics[width=\columnwidth]{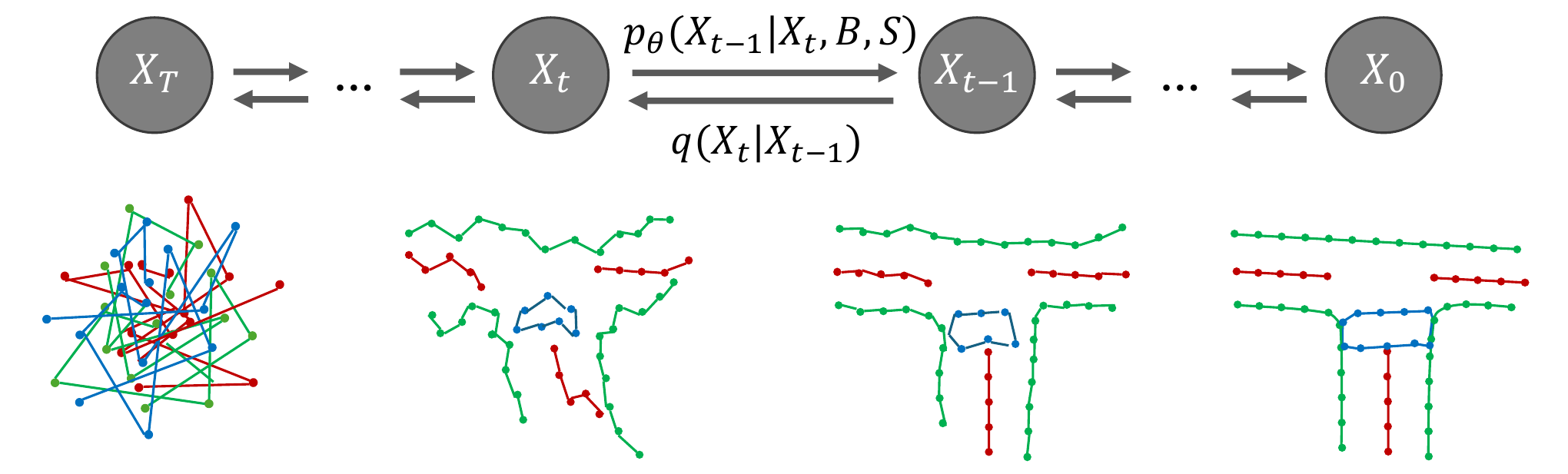}
    \caption{NavMapFusion diffusion process. The reverse process is conditioned on sensor data from $B$ and SD map data from $S$.}
    \label{fig:map_diffusion_process}
\end{figure}

\subsection{Problem Statement} \label{sec:problem}
Let $U=\{u_1, \ldots, u_n\}$ be the set of image frame sequences from the $n$ monocular cameras mounted on the ego vehicle.
Moreover, let $\mathcal{P}_{\operatorname{div}}$, $\mathcal{P}_{\operatorname{bound}}$, and $\mathcal{P}_{\operatorname{cross}}$ be the set of polylines (each polyline $P = \left[ (x_{i},y_{i} ) \right]_{i=1}^{N_{P}}$ is a sequence of points) representing lane dividers, lane boundaries, and pedestrian crossing within the scene, respectively,
$\mathcal{M}_{HD} = \{\mathcal{P}_{\operatorname{div}}, \mathcal{P}_{\operatorname{bound}}, \mathcal{P}_{\operatorname{cross}}\}$ be the local HD map with ego vehicle at the origin.
Let $\mathcal{M}_{SD}$ be the navigation map consisting of the set of polylines $\mathcal{P}_{\operatorname{road}}$ representing road geometries.
The goal is to find a function $m$ that returns the local HD map $\mathcal{M}_{HD}$ for a given sequence of sets of image frames ${U}$ and the navigation map $\mathcal{M}_{SD}$:

\begin{equation}
    \mathcal{M}_{HD} = m \left( U, \mathcal{M}_{SD} \right).
    \label{eq:problem_statement}
\end{equation}

\begin{figure*}[t]
    \centering
    \includegraphics[width=1.0\textwidth]{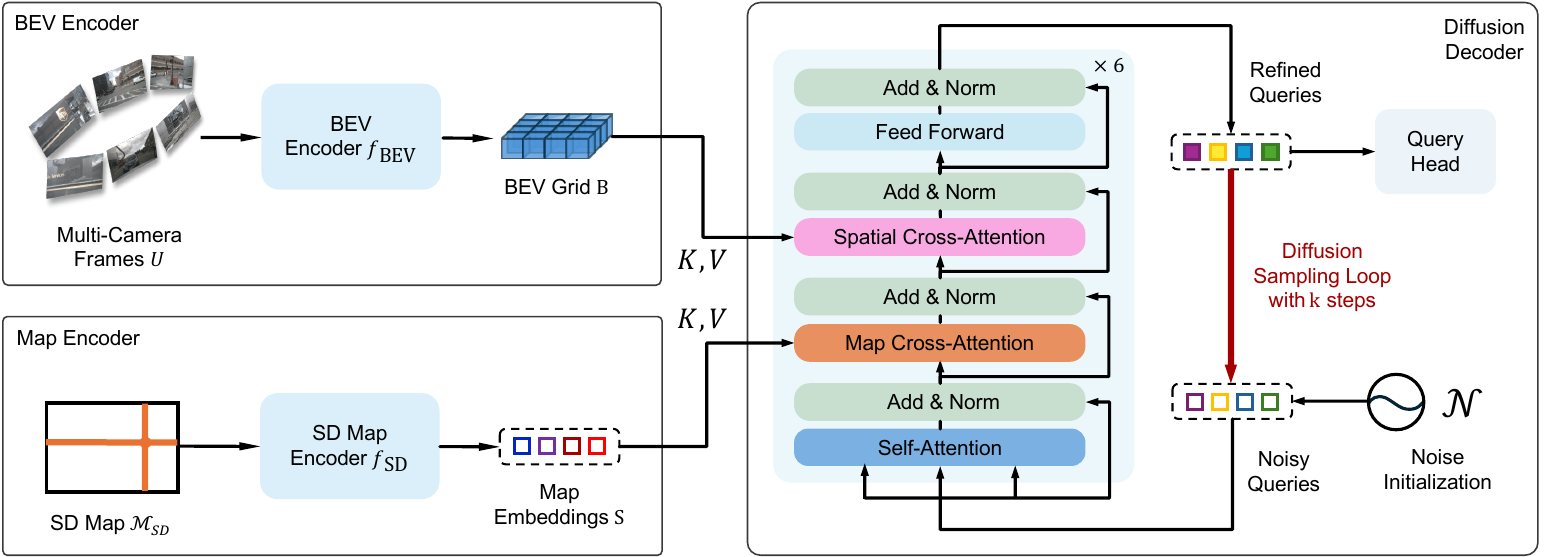}
    \caption{NavMapFusion Architecture with BEV Encoder $f_{BEV}$, SD Map Encoder $f_{SD}$, and Diffusion Decoder $g$.}
    \label{fig:navmap_diffusion_architecture}
\end{figure*}

\subsection{Diffusion for Map Construction}\label{sec:diffusion_map_construction}

Denoising Diffusion Probabilistic Models (DDPMs) \cite{ho2020denoising} generate data by learning to reverse a Markovian forward process that gradually adds Gaussian noise to \( {X}_{0} \), where $X_0$ is the vectorized GT map $\mathcal{M}_{HD}$. 
The forward process defines a noisy sample at timestep \( t \) as:
\begin{equation}
{X}_t = \sqrt{\bar{\alpha}_t} {X}_{0} + \sqrt{1 - \bar{\alpha}_t} \, \epsilon, \quad \epsilon \sim \mathcal{N}(0, \mathbf{I})
\end{equation}
where $\alpha_t = 1 - \beta_t$ and $\bar{\alpha}_t = \prod_{s=1}^{t}\alpha_s$ control the noise schedule based on hyperparameters $\{\beta_t\}_{t=1}^T$.
 
The reverse process iteratively denoises \( {X}_T \) back to \( {X}_{0} \).
Both processes are visualized for vectorized maps in \cref{fig:map_diffusion_process}.

A neural network with learnable weights $\theta$, \( \epsilon_\theta({X}_t, t) \), is trained to minimize the mean squared error between the true and predicted noise:
\begin{equation}
L(\theta) = \mathbb{E}_{{X}_{0}, t, \epsilon} \left[ \left\| \epsilon - \epsilon_\theta({X}_t, t) \right\|^2 \right].
\end{equation}

Once trained, the model can construct a map from pure Gaussian noise \( {X}_T \) by iteratively applying the reverse process:
\begin{equation}
{X}_{t-1} = \frac{1}{\sqrt{\alpha_t}} \left( {X}_t - \frac{1 - \alpha_t}{\sqrt{1 - \bar{\alpha}_t}} \epsilon_\theta({X}_t, t) \right) + \sigma_t z,
\end{equation}

where $z \sim \mathcal{N}(0, \mathbf{I})$ is Gaussian noise and $\sigma_t$ is a hyperparameter controlling the stochasticity of the reverse step.

\subsection{Conditional Diffusion-based Map Construction}\label{sec:diffusion_conditioning}

NavMapFusion uses conditioning to guide the diffusion-based map construction process introduced in \cref{sec:diffusion_map_construction}.
Specifically, to include features from the camera sensors, NavMapFusion adopts the conditioning from MapDiffusion \cite{monninger2025mapdiffusion}.
Spatial Cross-Attention (SCA) is performed with the map element queries $Q$ attending to features from the BEV grid $B$.
SCA is implemented with deformable attention \cite{zhu2021deformable}.
Formally, for a query feature $z_q$ located at reference point $v_q$ in a BEV grid $B\in\mathbb{R}^{H\times W\times C}$,
\begin{equation}
\operatorname{SCA}(z_q,v_q,B) = \sum_{k=1}^{K} A_{qk}\, W\,B\!\bigl[v_q+\Delta v_{qk}\bigr],
\label{eq:foundation:sca}
\end{equation}
where $A_{qk}$ and $\Delta v_{qk}$ are, respectively, the attention weight and offset of the $k$-th sampling point, and $W$ is a learnable weight matrix.
The offsets and weights are predicted from $z_q$ to keep computation and memory linear in spatial size.
SCA specifically uses the variant from StreamMapNet \cite{streammapnet_2024} with multi-point attention.

We propose additional guidance of the diffusion process through SD map features.
To this end, we extend the transformer decoder with a Map Cross-Attention (MCA) step that conditions the denoising on map embeddings.
In our model, the query for cross-attention is derived from the noisy ground-truth input \( {X}_t \).
The SD map embeddings \( S \) provide the key and value representations. We compute the projections:
\[
Q = {X}_t W_Q, \quad K = S W_K, \quad V = S W_V,
\]
and apply standard scaled dot-product attention:
\begin{equation}
\operatorname{MCA}(Q, K, V) = \text{softmax} \left( \frac{Q K^\top}{\sqrt{d}} \right) V.
\end{equation}
This allows the noisy map representation to attend to spatial priors from the SD map during the denoising process.

MCA and SCA provide different degrees of information. 
We use a regularization technique to encourage a more balanced fusion strategy.
Random dropout is applied to the BEV grid $B$ by setting $B$ to zero with probability $d_{BEV}$.
This prevents the model from over-relying on the sensor-derived features in $B$ and forces it to better leverage the complementary information from $S$, yielding a more robust online map $\mathcal{M}_{HD}$.

\subsection{NavMapFusion Architecture} \label{sec:architecture}

NavMapFusion follows the learned BEV encoder paradigm that decomposes function $m$ into an encoder $f_{BEV}$, that creates a BEV grid $B$, and a decoder $g$, that generates the map $\mathcal{M}_{HD}$ conditioned on $B$ and $\mathcal{M}_{SD}$.
This process is:
\begin{align}
    \mathcal{M}_{HD} &= g \left( f_{BEV} \left( U \right), \mathcal{M}_{SD} \right).
    \label{eq:encoder_decoder}
\end{align}

The full architecture of NavMapFusion is shown in \cref{fig:navmap_diffusion_architecture}.
Multi-camera frames $U$ are encoded into a latent grid $B$ with a learned BEV encoder $f_{BEV}$.
The NavMapFusion process uses random queries from $\mathcal{N}(0, 1)$ as a starting point.
Following MapDiffusion \cite{monninger2025mapdiffusion}, this denoising is conditioned on the camera images via SCA and on the SD map via MCA to guide the process.
The diffusion model is optimized so that the reverse process learns to denoise the random queries such that a query head can predict an HD map $\mathcal{M}_{HD}$ from the refined queries.

The novelty of the NavMapFusion architecture is in the integration of prior map information, for example from a navigation map $\mathcal{M}_{SD}$.
This navigation map is first encoded with an SD map encoder $f_{SD}$ into map embeddings $S$.
For each of the polylines $P \in \mathcal{P}_{\operatorname{road}}$, $f_{SD}$ creates an individual embedding.
The resulting set $S$ is used as Keys and Values in MCA.
Conditioning the denoising process on navigation maps serves as additional information for the map construction task to complement the sensor information at larger perception ranges and occlusions.

Each decoder layer is composed of a self-attention block, a Map Cross-Attention (MCA) block, a Spatial Cross-Attention (SCA) block, and a feed-forward network, each followed by add and norm operations. 
The full diffusion decoder consists of $L$ of these transformer decoder layers and a MLP-based query head to decode vectorized map representations.

\subsection{Sampling} \label{sec:sampling}

Sampling is the process of generating data points iteratively. The number of sampling steps \( k \) is distinct from \( T \), the total number of steps defined for the forward noising process. For efficient generation, \( k \ll T \) is typically used. 

We adopt the Denoising Diffusion Implicit Model (DDIM) formulation \cite{song2021denoising} to reduce the required number of diffusion steps while preserving the quality of map construction. 
The diffusion process is applied only to the decoder \( g \), while the conditioning remains deterministic. Specifically, both the BEV encoder \( f_{\text{BEV}} \) and SD map encoder \( f_{\text{SD}} \) are executed once and reused throughout the denoising process. This design enables efficient multi-step sampling with latency suitable for real-time applications. The decoder \( g \) refines the vectorized map output through \( k \) DDIM-based denoising steps.

\begin{figure*}[tbp]
    \centering
    \includegraphics[width=1.0\textwidth]{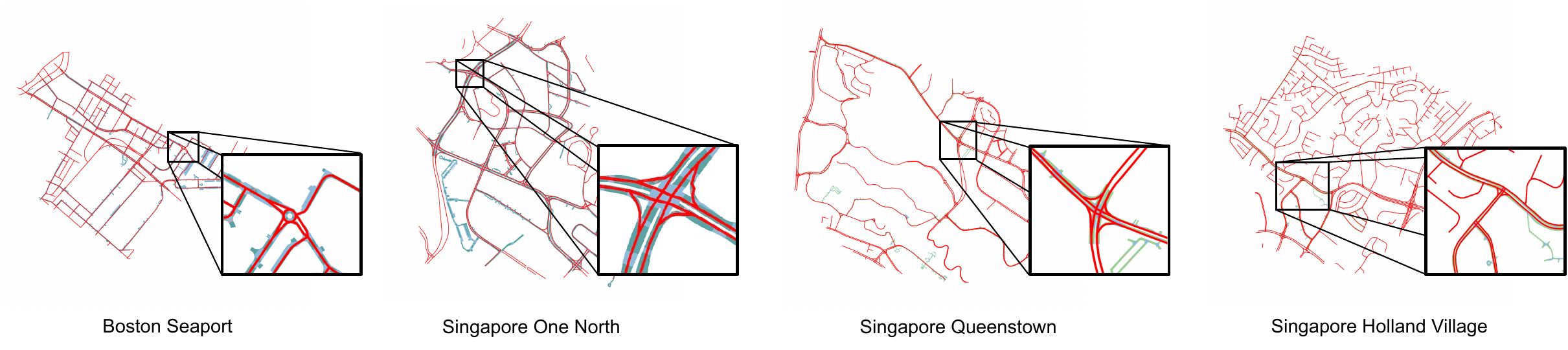}
    \caption{OpenStreetMap data used as $\mathcal{M}_{SD}$ visualized in red on top of nuScenes GT map $\mathcal{M}_{HD}$.}
    \label{fig:navmap_diffusion_sdmap}
\end{figure*}

\section{Experiments} \label{sec:experiments}

\subsection{Dataset and Evaluation Metrics}\label{sec:metrics}
We conduct our experiments on the NuScenes dataset \cite{nuscenes}, which provides data points at $\SI{2}{\hertz}$. Those include images from $n=6$ monocular cameras $U$ and corresponding vectorized GT maps $\mathcal{M}_{HD}$ that include elements from the categories road boundary ("bound"), lane divider ("div"), and pedestrian crossing ("ped"). 
We use the geospatially disjoint dataset splits from StreamMapNet \cite{streammapnet_2024}.
The performance on the vectorized map construction task is evaluated using mean Average Precision (mAP).

\subsection{Navigation Map Data} \label{sec:navigation_map_data}
The navigation map data from OpenStreetMap (OSM) is used as the prior map in the experiments.
We follow the code and data provided from P-MapNet \cite{jiang2024pmapnet} to pre-process the OSM data for the nuScenes dataset \cite{nuscenes}.
To simulate a scalable setting based on navigation map input, only road-level polylines $\mathcal{P}_{\operatorname{road}}$ are retained from the OSM. Manual alignment is performed for accurate localization of the prior map. 
\cref{fig:navmap_diffusion_sdmap} illustrates the OSM used as prior map \( \mathcal{M}_{SD} \), overlaid on the GT HD map \( \mathcal{M}_{HD} \) from nuScenes.

\subsection{Experimental Setup}\label{sec:implementation}
The training setup and the loss functions are taken from the reference architecture MapDiffusion \cite{monninger2025mapdiffusion}.
We adopt the training configuration of StreamMapNet with 24 epochs and batch size 1. 
The model training is performed in parallel on 8 NVIDIA V100 GPUs. 
AdamW is used for optimization with a cosine annealing schedule and a \num{2e-4} learning rate.
The size of the BEV grid is $100 \times 50$ with a default perception range of $\SI{100}{\meter} \times \SI{50}{\meter}$.

We use $f_{BEV}$ from the StreamMapNet model \cite{streammapnet_2024}.
For $f_{SD}$, the map encoder from SMERF \cite{luo2024smerf} is used.
The diffusion decoder uses $L=6$ refinement layers.
The dropout rate for SCA, \ie, for setting $B$ to zero, is $d_{BEV}=0.30$.

\subsection{Baseline Models}\label{sec:baseline}
MapDiffusion \cite{monninger2025mapdiffusion} is a diffusion-based approach that produces vectorized maps online directly from noise.
Hence, it is used as reference architecture and primary baseline to assess the improvement gained through using prior information from navigation maps.
StreamMapNet-MCA is a baseline that extends StreamMapNet \cite{streammapnet_2024} with $f_{SD}$ and MCA.
We create it to assess the benefit of performing MCA inside a diffusion framework \vs in a deterministic framework.
ScalableMap \cite{yu2023scalablemap} serves as baseline for long-range online map construction.
Baselines that use navigation maps are MapTR-SDMap \cite{jiang2024pmapnet}, P-MapNet \cite{jiang2024pmapnet}, MapEX \cite{sun2023mind}, and M3TR \cite{immel2024m3tr}.
Finally, we also compare against sensor-only baselines including VectorMapNet \cite{vectormapnet_2023}, MapTR \cite{maptr_2023}, StreamMapNet \cite{streammapnet_2024}, and SQD-MapNet \cite{wang2024stream}.

\subsection{Quantitative Results of NavMapFusion}\label{sec:quantitative_results}

{\begin{table}[tbp]
    \centering
    \caption{Performance of NavMapFusion compared to various baselines at perception range $\SI{100}{\meter} \times \SI{50}{\meter}$ on nuScenes split without geospatial overlap \cite{streammapnet_2024}. $^*$ show results from \cite{streammapnet_2024}, all other results are reproduced. AP thresholds $\{1.0, 1.5, 2.0\}$.}
    \label{tab:table_main_result_100x50}
    \resizebox{\columnwidth}{!}{%
    \begin{tabular}{l|cccc}
        Method & AP$_{\mathrm{ped}}$&  AP$_{\mathrm{div}}$  &  AP$_{\mathrm{bound}}$& mAP  \\ \midrule
        VectorMapNet$^*$ \cite{vectormapnet_2023} & 12.0 & 8.1 & 6.3 & 8.8 \\
        MapTR$^*$ \cite{maptr_2023} & 8.3 & 16.0 & 20.0 & 14.8 \\
        SQD-MapNet \cite{wang2024stream}  & 24.8 & 18.8 & 23.7 & 22.4 \\
        StreamMapNet \cite{streammapnet_2024} & 24.8  & 18.4 & 25.6 & 22.9 \\
        StreamMapNet-MCA (ours) & 28.4 & \textbf{22.2} & 27.8 & 26.1 \\
        MapDiffusion \cite{monninger2025mapdiffusion} & 23.4 & 21.6 & 22.2 & 22.4 \\
        NavMapFusion (ours) & \textbf{32.1} & 20.7 & \textbf{28.9} & \textbf{27.2} \\
    \end{tabular}%
    }
\end{table}
}
{\begin{table}[tbp]
    \centering
    \caption{Performance of NavMapFusion compared to baselines for long-range map construction.
    NavMapFusion is trained without dropout.
    $^\dagger$ range $\SI{120}{\meter}$, original nuScenes split \cite{nuscenes}, AP thresholds $\{1.0, 1.5, 2.0\}$.
    $^\ast$ range $\SI{120}{\meter}$, original nuScenes split \cite{nuscenes}, AP thresholds $\{0.5, 1.0, 1.5\}$.
    $^\ddagger$ range $\SI{100}{\meter}$, geospatially disjoint nuScenes split \cite{streammapnet_2024}, AP thresholds $\{1.0, 1.5, 2.0\}$. 
    }
    \label{tab:table_compare_prior}
    \resizebox{\columnwidth}{!}{%
    \begin{tabular}{l|cccc}
        Method & AP$_{\mathrm{ped}}$&  AP$_{\mathrm{div}}$  &  AP$_{\mathrm{bound}}$& mAP \\
        \midrule
        ScalableMap \cite{yu2023scalablemap}$^\dagger$ & 44.8 & 49.0 & 43.1 & 45.6 \\
        NavMapFusion (ours)$^\dagger$ & 57.5 & 60.2 & 57.3 & 58.4 \\
        \hline
        MapTR-SDMap \cite{maptr_2023,jiang2024pmapnet}$^\ast$ & 22.0 & 27.2 & 19.5 & 22.9 \\
        P-MapNet \cite{jiang2024pmapnet}$^\ast$ & 14.6 & 19.4 & 38.7 & 24.2 \\ 
        NavMapFusion (ours)$^\ast$ & 34.4 & 37.2 & 25.7 & 32.4 \\ 
        \midrule
        MapEX \cite{sun2023mind,immel2024m3tr}$^\ddagger$ & 22.0 & 17.2 & - & 19.6 \\
        M3TR \cite{immel2024m3tr}$^\ddagger$ & 24.7 & 18.9 & - & 21.7 \\ 
        NavMapFusion (ours)$^\ddagger$ & 28.1 & 22.8 & - & 25.5 \\
    \end{tabular}%
    }
\end{table}
}

\cref{tab:table_main_result_100x50} shows the results of NavMapFusion in comparison with online map construction baselines on perception range $\SI{100}{\meter} \times \SI{50}{\meter}$. 
Our core hypothesis is that a diffusion framework fuses coarse map priors more effectively than a deterministic one. 
The data validates this: while the deterministic StreamMapNet-MCA achieves a \SI{14.0}{\percent} relative gain over its baseline (from \SI{22.9}{\percent}~mAP to \SI{26.1}{\percent}~mAP), our diffusion-based NavMapFusion boosts mAP by \SI{21.4}{\percent} (from \SI{22.4}{\percent}~mAP to \SI{27.2}{\percent}~mAP). This means the relative improvement from diffusion is \SI{52.9}{\percent} higher than the deterministic case (\SI{+21.4}{\percent} \vs \SI{+14.0}{\percent}), confirming the significant benefit of fusing navigation map information through a generative process.

The class-specific results show that the improvement is most significant for road boundaries (\SI{+30.2}{\percent}) and pedestrian crossings (\SI{+37.2}{\percent}).
This is expected since the road polylines $\mathcal{P}_{\operatorname{road}}$ from $\mathcal{M}_{SD}$ provide road-level information, and crossing road lines suggest intersections that often come with pedestrian crossings.
Furthermore, NavMapFusion outperforms earlier methods such as VectorMapNet and MapTR, as well as state-of-the-art methods like StreamMapNet and SQD-MapNet.

\cref{tab:table_compare_prior} shows the results of NavMapFusion (trained without dropout) in comparison with baselines designed for long-range map construction either by specialized architectures or by using prior maps.
The first part of \cref{tab:table_compare_prior} matches the experimental setup of ScalableMap \cite{yu2023scalablemap}.
In that setup, NavMapFusion reaches \SI{58.4}{\percent}~mAP, exceeding the \SI{45.6}{\percent}~mAP achieved by ScalableMap.
Given that ScalableMap substantially outperforms SuperFusion \cite{dong2024superfusion}, our results demonstrate that NavMapFusion surpasses both long-range mapping methods with the help of prior maps.

The second part of \cref{tab:table_compare_prior} matches the experimental setup of P-MapNet \cite{jiang2024pmapnet}, which uses prior maps.
P-MapNet \cite{jiang2024pmapnet} reaches \SI{24.2}{\percent}~mAP.
MapTR-SDMap, which is presented as a baseline in P-MapNet \cite{jiang2024pmapnet}, reaches \SI{22.9}{\percent}~mAP.
NavMapFusion outperforms both baselines reaching \SI{32.4}{\percent}~mAP.

The third part of \cref{tab:table_compare_prior} shows results on the default experimental setup in this work.
Available baselines matching this setup are MapEX \cite{sun2023mind} and M3TR \cite{immel2024m3tr}.
A direct comparison is still difficult since both works use a subset of the GT map $\mathcal{M}_{HD}$ as prior information.
The variant integrating GT road boundaries is, while much more high-fidelity, the most similar to our conditioning.
In that variant, MapEX \cite{sun2023mind} and M3TR \cite{immel2024m3tr} both achieve close-to-perfect $AP_{\operatorname{bound}}$. 
We follow their evaluation protocol by calculating the mAP only on $AP_{\operatorname{ped}}$ and $AP_{\operatorname{div}}$.
With provisioning of GT road boundaries, the mAP performance of MapEX \cite{sun2023mind} on the remaining classes reaches \SI{19.6}{\percent}.
M3TR \cite{immel2024m3tr} slightly improves over MapEX with \SI{21.7}{\percent}~mAP.
Our NavMapFusion approach achieves an even larger mAP of \SI{25.5}{\percent}, again outperforming both baselines.
In summary, on both original and new nuScenes split, NavMapFusion achieves state-of-the-art on absolute performance and also yields strong relative improvements over its reference architecture thanks to effectively integrating information from $\mathcal{M}_{SD}$.

\cref{tab:table_range} shows the results of NavMapFusion (trained without dropout) on various perception ranges in comparison to MapDiffusion, the reference architecture with no MCA.
While the relative improvement is negligible at $\SI{60}{\meter} \times \SI{30}{\meter}$, it increases substantially with larger perception ranges.
NavMapFusion reaches \SI{4.8}{\percent} relative improvement on $\SI{80}{\meter}$ \vs \SI{18.8}{\percent} on $\SI{100}{\meter}$.
The relative improvement increases further with \SI{52.5}{\percent} on $\SI{120}{\meter}$ and \SI{57.4}{\percent} on $\SI{150}{\meter}$.
This immense improvement confirms the hypothesized benefit of using navigation maps as prior information for online map construction.
The benefit is higher for larger perception ranges since sensor limitations such as perception range and likelihood for occlusion become stronger influencing factors.
NavMapFusion proposes an effective way to combine coarse SD maps with high-fidelity image data, improving the map construction performance at larger ranges while maintaining it in near range.
It achieves this at 14.7~FPS for $k=1$ and at 8.1~FPS for $k=5$ on an Nvidia A6000 GPU, maintaining real-time capabilities.

{\begin{table}[tbp]
    \centering
    \caption{Comparison of NavMapFusion without MCA (\ie, MapDiffusion, \xmark) and NavMapFusion (\cmark) at multiple perception ranges on nuScenes split without geospatial overlap \cite{streammapnet_2024}. $d=0.0$. AP thresholds $^\ddagger$: \{0.5, 1.0, 1.5\}, $^\ast$: \{1.0, 1.5, 2.0\}.}
    \label{tab:table_range}
    \resizebox{\columnwidth}{!}{%
    \begin{tabular}{lc|cccc|r}
        Range & Map & AP$_{\mathrm{ped}}$&  AP$_{\mathrm{div}}$  &  AP$_{\mathrm{bound}}$& mAP   &Impr.\\ \midrule
        \multirow{2}{*}{$\SI{60}{\meter} \times \SI{30}{\meter}^\ddagger$} & \xmark & 32.9 & 31.4 & 42.4 & 35.6  &\\
        & \cmark & 31.8 & 30.7 & 44.4 & 35.6  &$\SI{+0.0}{\percent}$\\
        \midrule
        \multirow{2}{*}{$\SI{80}{\meter} \times \SI{40}{\meter}^\ddagger$} & 
        \xmark & 18.9 & 19.5 & 24.3 & 20.9  &\\
        & \cmark & 20.2 & 19.8 & 25.7 & 21.9  &$\SI{+4.8}{\percent}$\\
        \midrule
        \multirow{2}{*}{$\SI{100}{\meter} \times \SI{50}{\meter}^\ast$} & \xmark & 23.4 & 21.6 & 22.2 & 22.4  &\\
        & \cmark & 28.1 & 22.8 & 29.0 & 26.6  &$\SI{+18.8}{\percent}$\\
        \midrule
        \multirow{2}{*}{$\SI{120}{\meter} \times \SI{60}{\meter}^\ast$} & 
        \xmark & 17.2 & 7.66 & 10.4 & 11.8  &\\
        & \cmark & 22.4 & 14.0 & 17.7 & 18.0  &$\SI{+52.5}{\percent}$\\
        \midrule
        \multirow{2}{*}{$\SI{150}{\meter} \times \SI{75}{\meter}^\ast$} & 
        \xmark & 11.4 & 4.2& 2.9& 6.1&\\
        & \cmark & 13.0 & 8.2 & 7.7 & 9.6  &$\SI{+57.4}{\percent}$\\
    \end{tabular}%
    }
\end{table}}
{\begin{table}
    \centering
    \caption{Ablation on diffusion parameters $k$, $\eta$, $\tau$ at perception range $\SI{100}{\meter} \times \SI{50}{\meter}$ on nuScenes split without geospatial overlap \cite{streammapnet_2024}. $d=0.0$. AP thresholds $\{1.0, 1.5, 2.0\}$.}
	\label{tab:table_diffusion_parameters}
    \resizebox{\columnwidth}{!}{%
    \begin{tabular}{cllccccc}
        Steps $k$ & $\eta$ & $\tau$ & FPS & AP$_{\mathrm{ped}}$&  AP$_{\mathrm{div}}$  &  AP$_{\mathrm{bound}}$& mAP \\
        \midrule
        1 & 0.5 & 0.5 & 14.7 & 27.7 & 22.6 & 28.9 & 26.4 \\
        3 & 0.5 & 0.5 & 10.3 & 28.1 & 22.7 & 28.9 & 26.6 \\
        5 & 0.5 & 0.5 & 8.1 & 28.1 & 22.8 & 29.0 & 26.6 \\
        \midrule
        5 & 0.1 & 0.5 & 8.1 & 28.0 & 22.7 & 28.9 & 26.5 \\
        5 & 0.5 & 0.5 & 8.1 & 28.1 & 22.8 & 29.0 & 26.6 \\
        5 & 0.9 & 0.5 & 8.1 & 28.1 & 22.8 & 29.0 & 26.6 \\
        \midrule
        5 & 0.5 & 0.1 & 8.1 & 27.5 & 22.1 & 28.7 & 26.1 \\
        5 & 0.5 & 0.5 & 8.1 & 28.1 & 22.8 & 29.0 & 26.6 \\
        5 & 0.5 & 0.9 & 8.1 & 28.1 & 22.9 & 29.0 & 26.6 \\
    \end{tabular}%
    }
\end{table}}
{\begin{table}[tbp]
    \centering
    \caption{Ablation of NavMapFusion at perception range $\SI{100}{\meter} \times \SI{50}{\meter}$ on nuScenes split without geospatial overlap \cite{streammapnet_2024}. ``Impr.'' is the \emph{incremental rel.} improvement. AP thresholds $\{1.0, 1.5, 2.0\}$.}
    \label{tab:table_ablation}
    \resizebox{\columnwidth}{!}{%
    \begin{tabular}{l|cccc|r}
        Method & AP$_{\mathrm{ped}}$&  AP$_{\mathrm{div}}$  &  AP$_{\mathrm{bound}}$& mAP & Impr. \\ 
        \midrule
        Baseline \cite{monninger2025mapdiffusion} & 23.4 & 21.6 & 22.2 & 22.4 &  \\
        + MLP-based $f_{SD}$ & 26.3 & 20.7 & 28.3 & 25.1 & $\SI{+12.1}{\percent}$ \\
        + SMERF \cite{luo2024smerf}-based $f_{SD}$ & 31.5 & 18.9 & 28.6 & 26.3 & $\SI{+4.8}{\percent}$  \\
        + MCA before SCA & 28.1 & 22.8 & 29.0 & 26.6 & $\SI{+1.1}{\percent}$ \\
        + Dropout on $B$ (ours) & 32.1 & 20.7 & 28.9 & 27.2 & $\SI{+2.3}{\percent}$ \\
    \end{tabular}%
    }
\end{table}}

\begin{figure}[tbp]
    \centering
    \includegraphics[width=\columnwidth]{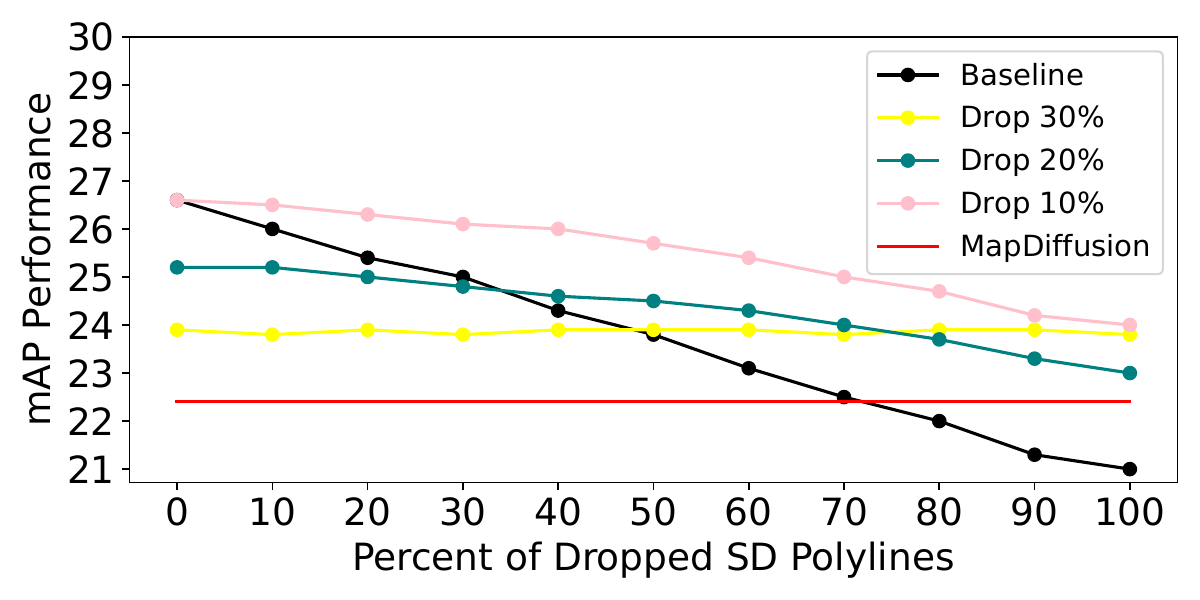}
    \vspace{-7mm}
    \caption{Robustness to outdated map geometries. mAP performance \vs percentage of SD polylines dropped at test-time. Lines show models trained with different dropout rates.}
    \label{fig:robustness_analysis}
\end{figure}
\begin{figure}[tbp]
    \centering
    \includegraphics[width=\columnwidth]{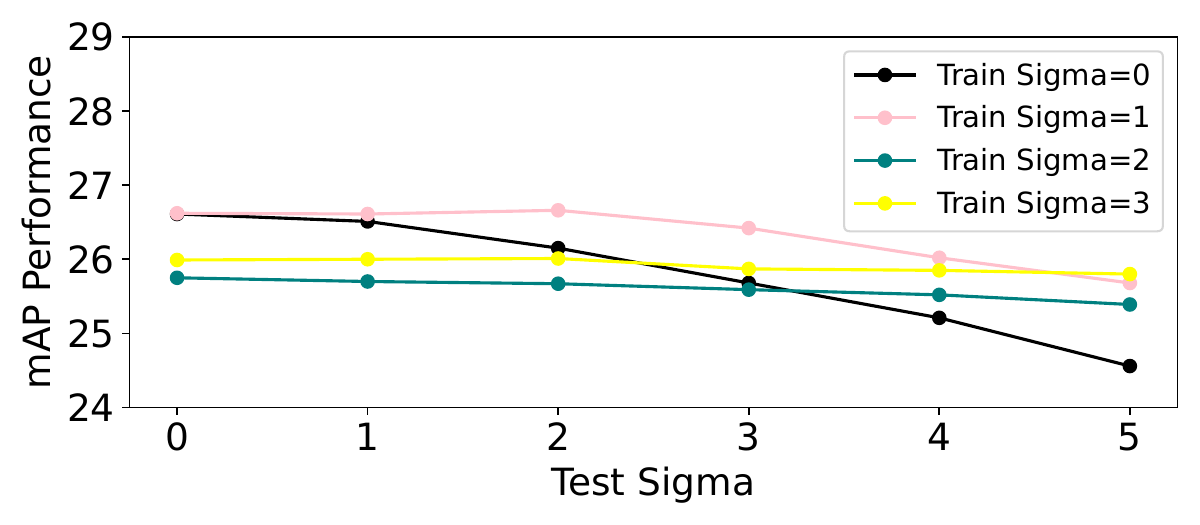}
    \vspace{-7mm}
    \caption{Robustness to location inaccuracy. mAP performance \vs standard deviation ($\sigma$ in meters) of Gaussian noise applied at test-time. Lines show models trained with different noise levels $\sigma$.}
    \label{fig:random_noise}
\end{figure}

\subsection{Ablation Studies}\label{sec:Ablation Studies}

\cref{tab:table_diffusion_parameters} shows the performance of NavMapFusion trained without dropout and evaluated on various key diffusion parameters, including the number of diffusion steps $k$, the $\eta$ parameter in DDIM sampling \cite{song2021denoising}, and the query threshold~$\tau$.
For $k$, the map construction quality increases with more steps and saturates at around $k=5$.
The numbers show a favorable performance \vs latency trade-off.
Besides for performance, multiple diffusion steps are important for generating diverse samples as explored in MapDiffusion \cite{monninger2025mapdiffusion}.
The $\eta$ parameter controls the randomness in the generation process.
Since map construction is less multi-modal, this parameter has only a minor influence, with the best performance given between $\eta=0.5$ and $0.9$.
The query threshold $\tau$ shows a similar trend with the best performance between $\tau=0.5$ and $0.9$.
For the final evaluation, NavMapFusion uses $k=5$, $\eta=0.5$, and $\tau=0.5$.

\begin{figure*}
    \centering
    \begin{subfigure}[t]{1\textwidth}
        \centering
        \includegraphics[trim=0cm 0cm 0cm 0cm, clip, width=\textwidth]{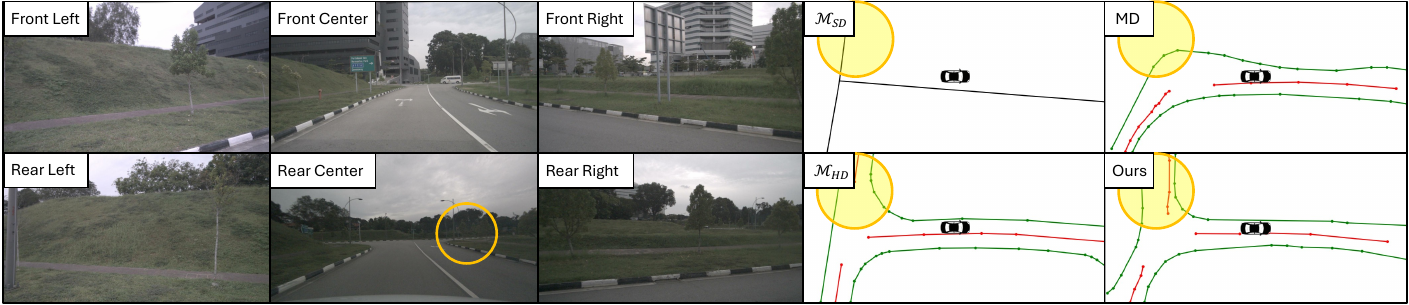}
        \caption{Intersection behind ego vehicle barely visible. MapDiffusion misses one road (yellow circle) that is predicted by NavMapFusion thanks to $\mathcal{M}_{SD}$.}
        \label{fig:qualitative_result_1}
    \end{subfigure}
    \\
    \begin{subfigure}[t]{1\textwidth}
        \centering
        \includegraphics[trim=0cm 0cm 0cm 0cm, clip, width=\textwidth]{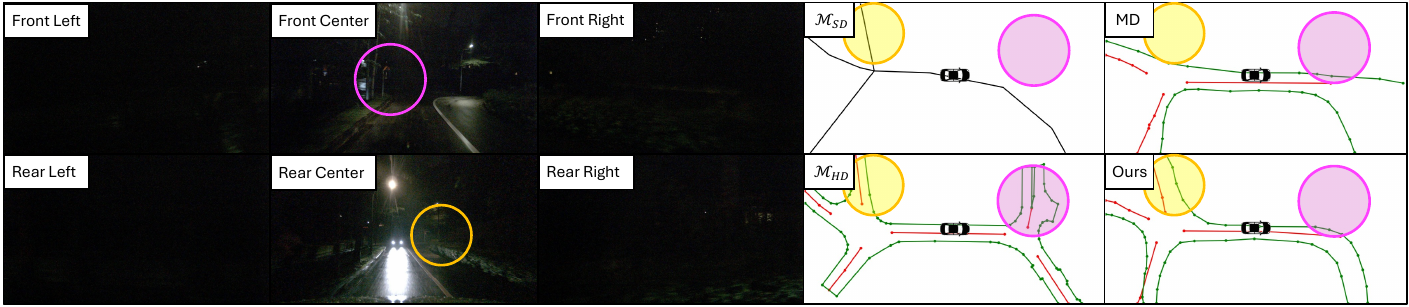}
        \caption{Night scene with poor illumination. Intersections in front (pink circle) and behind the ego vehicle (yellow circle) are barely visible. MapDiffusion misses both roads. NavMapFusion predicts the road behind ego correctly, but omits the upcoming left turning road, because it is missing in $\mathcal{M}_{SD}$.}
        \label{fig:qualitative_result_2}
    \end{subfigure}
    \vspace{-1mm}
    \caption{Two qualitative results of NavMapFusion. Camera images $U$ are on the left. Next to it are SD map $\mathcal{M}_{SD}$ and GT HD map $\mathcal{M}_{HD}$. On the right side are prediction of the MapDiffusion baseline without prior map (``MD'') and prediction of NavMapFusion (``Ours'').} 
    \label{fig:qualitative_result}
\end{figure*}

\cref{tab:table_ablation} shows an ablation of the NavMapFusion architecture.
Individual design choices contribute to the overall performance, starting from the baseline MapDiffusion \cite{monninger2025mapdiffusion} with \SI{22.4}{\percent}~mAP.
Integrating navigation map information by creating map embeddings $S$ with a 4-layer MLP for $f_{SD}$ improves the sensor-only performance by \SI{12.1}{\percent}, confirming that this diffusion framework can effectively leverage even simple map embeddings.
Using the sophisticated transformer-based map encoder from SMERF \cite{luo2024smerf} for $f_{SD}$ adds another relative improvement of \SI{4.8}{\percent}.
Finally, applying MCA before instead of after SCA increases relative performance by an additional \SI{1.1}{\percent}.
This confirms the theory from a Bayesian standpoint: MCA provides the coarse prior, and SCA subsequently updates this prior with evidence from sensor data.
Finally, adding dropout by setting $B$ to zero with probability $d_{BEV}$ adds another \SI{2.3}{\percent} relative improvement as rationalized in \cref{sec:diffusion_conditioning}, leading to a final NavMapFusion model with \SI{27.2}{\percent}~mAP.

\subsection{Robustness towards Imperfect SD Map Prior}\label{sec:robustness}
Our qualitative analysis (\cref{fig:qualitative_result_2}) shows NavMapFusion can be overly-reliant on the prior, omitting road elements visible to sensors but missing from $\mathcal{M}_{SD}$. This motivates a detailed study of the model's sensitivity to various failure modes that lead to imperfect SD map priors, for which we also investigate methods to improve robustness.

\subsubsection{Outdated Map Geometries}\label{sec:sd_dropout}
We simulate missing map geometries by randomly dropping SD map polylines from $S$ with probability $d_{SD}$.
The performance for various $d_{SD}$ is shown in \cref{fig:robustness_analysis}, with the red line corresponding to inference without any SD map condition, serving as a lower-bound reference. 
The baseline, NavMapFusion (black), exhibits a performance drop when evaluated with incomplete SD maps; with mAP values even below the lower bound with no SD map condition for $d_{SD}>0.7$ during inference. 
To improve robustness, we experiment with adding train-time dropout, resulting in a more gradual decline of performance.
Interestingly, higher train-time dropout rates introduce a trade-off: \SI{20}{\percent} dropout (green) achieves higher performance under ideal conditions, while \SI{30}{\percent} (yellow) offers more stable robustness for larger dropout rates. 
Notably, \SI{10}{\percent} dropout (pink) is optimal along the entire range. It enhances robustness at higher dropout rates but also maintains competitive performance when the full map prior is available. 

\subsubsection{SD Map Misalignment}\label{sec:sd_alignment}
We evaluate the importance of our manual alignment process of the SD maps (\cref{sec:navigation_map_data}).
For NavMapFusion trained on aligned SD maps, we get \SI{26.6}{\percent}~mAP when evaluating on aligned SD maps, and \SI{26.0}{\percent}~mAP on unaligned SD maps.
NavMapFusion trained on unaligned SD maps reaches \SI{26.8}{\percent}~mAP when evaluating on aligned SD maps, and \SI{26.5}{\percent}~mAP on unaligned SD maps.
In summary, alignment has a marginal impact, and using unaligned SD maps during training even improves testing on aligned SD maps, likely serving as a regularization.

\subsubsection{Location Inaccuracy}\label{sec:location_noise}
Inspired by the previous result, we evaluate NavMapFusion on translation errors sampled from random Gaussian noise with $\sigma$ in meters.
\cref{fig:random_noise} shows evaluations on random noise with different standard deviations, black is the model trained with no noise.
While reaching \SI{26.6}{\percent}~mAP for evaluation with no noise ($\sigma=0$), performance drops quite a bit for $\sigma > 1$.
As per expectation, stronger noise in training makes NavMapFusion more invariant to the navigation map input in general: Training with $\sigma > 1$ (green, yellow) is more robust to noise during evaluation, but also subpar in a setting without noise.
We find training with $\sigma = 1$ (pink) to be a good compromise, preserving maximum performance that is maintained up to evaluation noise $\sigma = 3$.

\subsubsection{Localization Errors}\label{sec:localization_errors}
For further analysis, we simulate two kinds of localization errors.
Using SD maps from a random location within the map area (ego potentially not near a road) with probability $p$, NavMapFusion drops from \SI{26.6}{\percent}~mAP for $p=0.0$ to \SI{20.1}{\percent}~mAP for $p=1.0$.
We inject these localization errors during training with $p=0.1$ to address robustness. NavMapFusion reaches evaluation results of \SI{25.9}{\percent}~mAP for $p=0.0$ and \SI{24.4}{\percent}~mAP for $p=1.0$, showing mostly preserved performance and much improved robustness.

More challenging to disambiguate are random valid locations, for example erroneously using the SD map from a past location. 
To simulate this, we use the localized SD map of a random other nuScenes data point with probability $p$.
NavMapFusion drops from \SI{26.6}{\percent}~mAP for $p=0.0$ to \SI{17.8}{\percent}~mAP for $p=1.0$ (even lower than aforementioned \SI{20.1}{\percent}~mAP for purely random locations).
Injecting these localization errors during training with $p=0.1$ gives evaluation results of \SI{24.7}{\percent}~mAP for $p=0.0$ and \SI{24.0}{\percent}~mAP for $p=1.0$.
This indicates high fault tolerance while still outperforming the baseline MapDiffusion (\SI{22.4}{\percent}~mAP).

\subsection{Qualitative Results}\label{sec:qualitative_results}
\cref{fig:qualitative_result} shows qualitative results of our NavMapFusion model in comparison with the MapDiffusion baseline on two traffic scenes.
In both scenes, NavMapFusion predicts a road branch missed by MapDiffusion thanks to leveraging additional information from $\mathcal{M}_{SD}$. 
In \cref{fig:qualitative_result_2}, additionally, MapDiffusion misses an upcoming road branch on the front left. That road branch is also omitted by NavMapFusion because it is missed in $\mathcal{M}_{SD}$.
This shows that while NavMapFusion can effectively fuse information from sensor data and prior map, it is limited to either of the inputs providing the information. 

\section{Conclusion} \label{sec:conclusion}
We introduced NavMapFusion, a novel diffusion-based framework for online vectorized HD map construction. 
The model learns to iteratively denoise random initializations under the guidance of sensor data and prior map. 
By conditioning the map constructing task individually on sensor data and map data, NavMapFusion effectively fuses low-fidelity prior information with high-fidelity sensor inputs.
NavMapFusion uniquely interprets discrepancies between the navigation prior and online sensor observations as noise within the diffusion framework. 
Our experiments demonstrate that NavMapFusion leverages prior map information more effectively than deterministic baselines, while maintaining real-time speed.
A detailed robustness study indicates that train-time regularization can increase robustness towards errors in the SD map while leveraging its benefits when the SD map is correct.
The benefit increases with larger perception ranges, confirming that prior maps are specifically helpful to complement sensor limitations.
This improves the robustness of downstream planning tasks, leading to safer autonomous driving.

{
    \small
    \bibliographystyle{ieeenat_fullname}
    \bibliography{main}
}

\end{document}